\documentclass[letterpaper]{article}
\usepackage{times}
\usepackage{helvet}
\usepackage{hyperref}
\usepackage[final, nonatbib]{tackling_climate_workshop_style}
\usepackage{graphicx}
\usepackage[style=numeric-comp]{biblatex}
\usepackage{bm}
\usepackage{appendix}
\usepackage{array}
\usepackage{booktabs}

\title{Difference Learning for Air Quality Forecasting Transport Emulation}

\author{%
  Reed R.~Chen\\
  Johns Hopkins Applied Physics Laboratory\\
  Laurel, MD 20723 \\
  \texttt{reed.chen@jhuapl.edu} \\
  \And
  Christopher Ribaudo \\
  Johns Hopkins Applied Physics Laboratory \\
  Laurel, MD 20723 \\
  \texttt{chris.ribaudo@jhuapl.edu} \\
  \AND
  Jennifer Sleeman \\
  Johns Hopkins Applied Physics Laboratory \\
  Laurel, MD 20723 \\
  \texttt{jennifer.sleeman@jhuapl.edu} \\
  \And
  Chace Ashcraft \\
  Johns Hopkins Applied Physics Laboratory \\
  Laurel, MD 20723 \\
  \texttt{chace.ashcraft@jhuapl.edu} \\
  \And
  Collin Kofroth \\
  Johns Hopkins Applied Physics Laboratory \\
  Laurel, MD 20723 \\
  \texttt{collin.kofroth@jhuapl.edu} \\
  \And
  Marisa Hughes\\
  Johns Hopkins Applied Physics Laboratory \\
  Laurel, MD 20723 \\
  \texttt{marisa.hughes@jhuapl.edu} \\
  \And
  Ivanka Stajner \\
  NOAA \\
  College Park, MD 20740 \\
  \texttt{ivanka.stajner@noaa.gov} \\
    \And
  Kevin Viner \\
  NOAA \\
  College Park, MD 20740 \\
  \texttt{kevin.viner@noaa.gov} \\
  \And
  Kai Wang \\
  NOAA \\
  College Park, MD 20740 \\
  \texttt{kai.wang@noaa.gov} \\
}

\begin{document}

\maketitle

\begin{abstract}
Human health is negatively impacted by poor air quality including increased risk for respiratory and cardiovascular disease. Due to a recent increase in extreme air quality events, both globally and locally in the United States, finer resolution air quality forecasting guidance is needed to effectively adapt to these events. The National Oceanic and Atmospheric Administration provides air quality forecasting guidance for the Continental United States. Their air quality forecasting model is based on a 15 km spatial resolution; however, the goal is to reach a three km spatial resolution. This is currently not feasible due in part to prohibitive computational requirements for modeling the transport of chemical species. In this work, we describe a deep learning transport emulator that is able to reduce computations while maintaining skill comparable with the existing numerical model. We show how this method maintains skill in the presence of extreme air quality events, making it a potential candidate for operational use. We also explore evaluating how well this model maintains the physical properties of the modeled transport for a given set of species.
\end{abstract}

\section{Introduction}
There has been a significant increase in high pollution air quality (AQ) events. These events are shown to have a strong sensitivity to extreme meteorological events such as heat waves \cite{hou2016long}. Increased wildfire activity has specifically contributed to a sudden increase in fine particulate matter (PM2.5) AQ pollution in the United States \cite{jaffe2020wildfire}. Studies show that increased emissions and climate change can negatively impact air quality \cite{moghani2020impact}. Since AQ has a direct correlation with increases in human-related illness and mortality \cite{ling2016global,landrigan2017air}, it is important for AQ forecasting to address this changing environment. Operational AQ forecasting guidance is provided by the National Oceanic and Atmospheric Administration (NOAA) for the Continental United States (CONUS). The NOAA forecasting guidance system is computationally challenged by the transport of chemical species which involves solving a set of physical governing equations. This transport is a critical component of modeling AQ and presents a challenge for reaching finer spatial resolutions. In this study, we explore the feasibility of using deep learning as a transport emulator to provide a speed-up in overall computation, potentially enabling finer resolution modeling. We built our method to tolerate bursts of extreme AQ events without a loss in skill. This is an important factor, as extreme AQ events present a challenge for the existing NOAA AQ forecasting system.

\subsection{Background}
NOAA provides operational forecast guidance for AQ, including ozone and PM2.5. NOAA uses the Unified Forecast System Air Quality (UFS-AQ) model for AQ forecasting \cite{stajner2023development}. The UFS-AQ transport of chemical tracers, which is used to calculate how species advect across the United States, is essential for AQ forecasts. Approximately 40\% of the overall computation time is spent in the transport module, where, for each grid point across CONUS, for each of the 64 vertical levels that represent vertical atmospheric conditions, and at each timestep, calculations must be performed. The sheer number of computations contributes to the intractability of reaching a three km resolution. 
The transport of 183 chemical tracers is used to provide 72-hour forecasts. Each of the chemical species is sequentially passed into the transport module approximately every 30 minutes. After the transport module, each of the species is also processed in two other processing modules, the physics module and the chemistry module (which is applied to all species simultaneously) before the next transport timestep. 

\subsection{Related Work}
Early Machine Learning (ML) methods which were used to speed up calculations by emulating parameterizations of atmospheric physics and chemistry include \cite{liao2020deep, kelp2018orders, shen2022machine}. Work related to chemistry emulation in an effort to replace this component of a model \cite{kelp2018orders} showed early promise of applying shallow neural networks to this class of problems. This work and the early promise of AI methods have been cautiously explored for weather forecasting. Data-driven AI methods have been shown to capture the physics of weather by using deep levels of abstraction across multiple layers of convolution. Recent research has shown impressive results when using data-driven AI methods for weather forecasting \cite{scher2018toward, weyn2020improving, rasp2020purely, rasp2021data} devoid of any explicit knowledge pertaining to the underlying physics. Many of the current state of the art weather forecasting AI models have shown excellent results forecasting a selection of weather variables and modeling a subset of vertical levels closest to the Earth's surface \cite{pathak2022fourcastnet,graphcast}. While models such as GraphCast \cite{graphcast} apply standard normalization to the residuals of each atmospheric variable, our model is species-agnostic, enabling a more flexible approach. Building on these ideas and our group's previous work \cite{sleeman2021integration, sleeman2023artificial} to emulate transport, we present a method that overcomes many issues present when working with atmospheric variables pertaining to chemical species such as highly skewed concentration distributions.

\section{Air Quality Data}
The NOAA AQ data was generated from the NOAA's AQM.v7.0 UFS-AQ model. The model was run for one month, and data pairs were captured before and after transport. The data pairs consisted of chemical species concentrations and meteorological variables across CONUS. Data was collected for seven days between September 1, 2020 and October 1, 2020 every five days (approximately five TBs of data), a period of time when active wildfires were present. The resolution of the 183 species at each time point was 232$\times$396$\times$64 pixels in latitude, longitude, and vertical levels. These data pairs were used to train the ML model. 

Data was subset to only include the 87 of the 183 species which contribute to the Air Quality Index (AQI) \cite{aqi}. These include the constituent species of PM2.5, O\textsubscript{3}, CO, NO\textsubscript{2}, SO\textsubscript{2}, and other contributors to the formation of ozone and PM2.5. For each species, data across CONUS was evenly divided into a 4$\times$6 grid of non-overlapping 58$\times$66$\times$16 pixel patches, and only the lowest 16 vertical levels, which most directly affect human health, were included. Since transport is advection-driven, the 3D wind velocity field and the altitude between vertical layers (which is not uniform) were included as features to the model. The surface geopotential, temperature, and pressure were also provided.

To assess the model's performance during extreme AQ events, patches were designated as either extreme or non-extreme based on the U.S. EPA's AQI thresholds for ozone and PM2.5. We considered AQI categories of \emph{moderate} or above as extreme \cite{aqi}. Thus, patches of ozone and ozone contributors were categorized as extreme if the maximum ozone concentration in the patch matched or exceeded 0.055 ppm. Similarly, patches of PM2.5 constituents were categorized as extreme if the maximum total PM2.5 concentration in that patch matched or exceeded 12.1 $\mu$g/m\textsuperscript{3}.

\subsection{Advection}
Modeling atmospheric variables and chemical species is difficult due to their highly skewed concentration distributions which vary greatly between variables and species. Traditionally, machine learning weather models normalize on a per-species basis, often with standard normalization. However, our model is species-agnostic and is not required to learn species-specific normalization parameters. Fortunately, the advection continuity equation, which the transport module solves for and our model aims to learn, is both species-agnostic and scale-invariant:
\[\frac{\partial c}{\partial t} + \nabla \cdot (\bm{v} c) = 0 \quad \Longleftrightarrow \quad \frac{\partial (ac)}{\partial t} + \nabla \cdot (\bm{v} ac) = 0,\] 
where \(c\) is the species concentration, \(\bm{v}\) is the velocity vector field, and \(a\) is the scaling factor. Furthermore, the advection equation is invariant under affine transformations if an incompressible flow is assumed due to the relatively low wind velocities: 
\[\frac{\partial c}{\partial t} + \nabla \cdot (\bm{v} c) = 0 \quad \Longleftrightarrow \quad \frac{\partial c}{\partial t} + \bm{v}\cdot \nabla  c = 0 \quad \Longleftrightarrow \quad \frac{\partial (ac + b)}{\partial t} + \bm{v} \cdot \nabla (ac + b) = 0,\] 
where \(a\) and \(b\) parameterize an affine transformation. As a result of this invariance, we are able to apply linear and affine transformations such as min-max normalization on a per-species, per-batch, or even per-patch basis while largely preserving the underlying transport. This enables extreme species concentrations that are spatio-temporally localized to specific patches, e.g. high PM2.5 concentrations caused by wildfires, to be normalized independently from other patches. In longer-range forecasting applications, this invariance can be useful for data distribution shifts caused by climate change and applying data normalization in continual learning settings. Additionally, it opens the possibility of applying log transforms to the highly right-skewed species concentrations, i.e. $c'=\ln(ac+b)$. For example, data can be min-max normalized to a range of 1 to \(e\) before the log transform to avoid zero or negative values. The machine learning model can then learn a governing equation analogous to
\[\frac{\partial \exp(c')}{\partial t} + \bm{v} \cdot \nabla \exp(c') = 0\] 
while maintaining invariance to species, batch, or patch-specific normalization parameters.

In our experiment, we demonstrated the potential of this approach by min-max normalizing the input and output data of the UFS-AQ model to a range of zero to one on a per-species basis. Since species concentrations decrease significantly as vertical level increases, we also min-max normalized by vertical layer. Although the advection equation is not invariant to this normalization, our approach aims to emulate rather than replicate the underlying transport.

\section{Methodology}
\subsection{Difference Learning}
\begin{figure}[ht]
    \centering
    \includegraphics[width=1\textwidth]{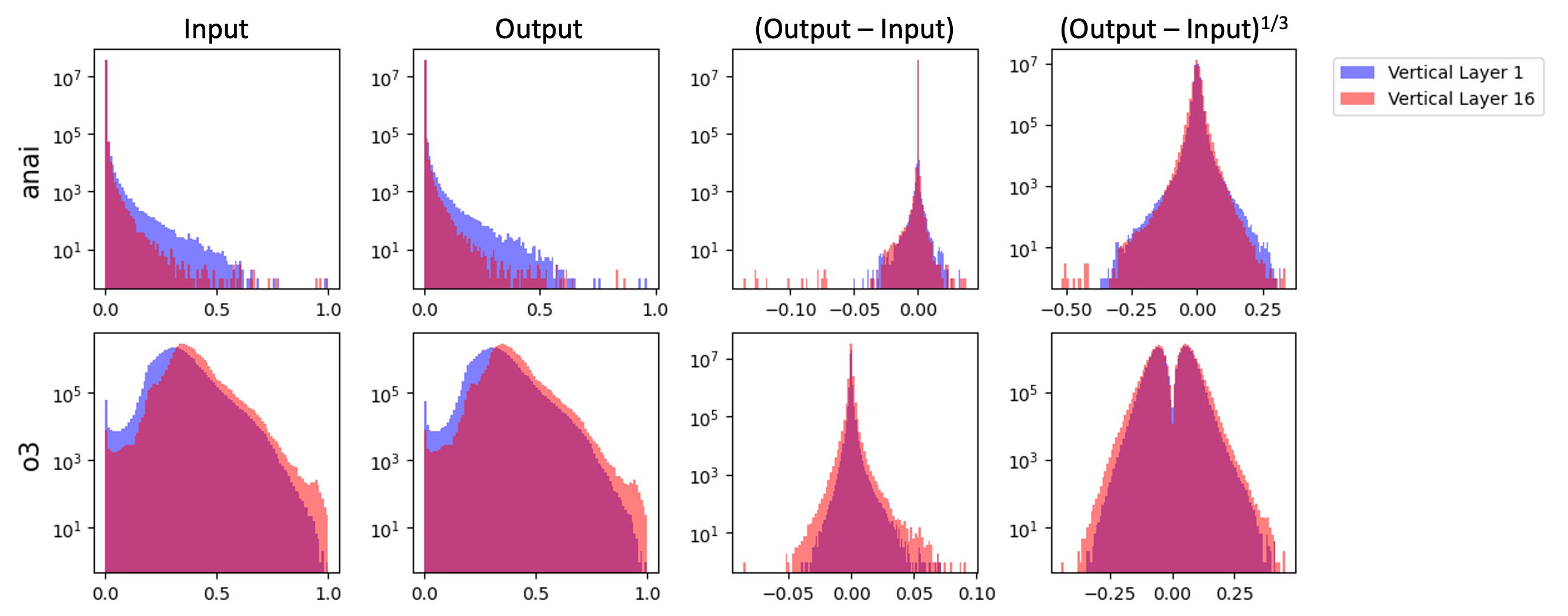}
    \caption{Distributions of chemical species \emph{anai}, a PM2.5 constituent, and ozone at vertical levels 1 (blue) and 16 (red). The $Input$ and $Output$ columns are concentration distributions from the UFS-AQ model after min-max normalization. Column 3 is the distribution of the difference between the $Input$ and $Output$ data. Column 4 is the distribution after taking the cube root of this difference. Note that the y-axis is log-scaled.}
    \label{fig:Distributions}
\end{figure}

Since the output of the UFS-AQ transport module is processed separately by the chemistry and physics modules at each timestep, our model must learn per-timestep advective transport. However, because the changes in species concentrations between the input and output of the UFS-AQ are relatively small as seen in Figure~\ref{fig:Distributions} (column 3), attempting to directly learn the translation between input and output can be akin to learning an autoencoding of the input data. Learning the small residuals between the output and input can also be challenging due to factors such as vanishing gradients. To overcome these challenges, we applied a cube root transformation to the difference. 

The cube root transformation is traditionally used to reduce the skewness of a distribution \cite{sinimaa2021feature}. For our data, it serves to increase the spread of the concentration distribution (Figure~\ref{fig:Distributions}). The cube root transformation has several immediate advantages; it does not require species-specific normalization parameters, can be applied to zero and negative values, and reshapes the target distribution to a wider range of -1 to 1. Since $n$'th root transformations are less sensitive to changes at the extremities of the range from -1 to 1, higher root transformations may yield more accurate predictions when the residuals are small, but may negatively impact prediction accuracy and contrast when the residuals are closer to -1 or 1 (Appendix, Figure~\ref{fig:root}). 

\subsection{Model and Training}
The deep learning model used to emulate the transport is a 3D U-Net with four downsampling and upsampling blocks \cite{ronneberger2015u}. The U-Net difference learning approach is illustrated in Figure~\ref{fig:diagram} (Appendix). The mean squared error (MSE) loss function and the Adam optimizer with a learning rate of 0.001, \(\beta_1 = 0.9\), and \(\beta_2 = 0.999\) were used. During training, the first 4 days of the data were divided into an 80/20 train/validation split, giving 394,214 training patches and 98,554 validation patches. The test dataset consisted of the last three days of data, or 369,576 patches. Of these 369,576 patches, 233,950 patches were classified as extreme. The U-Net has a total of 90,310,657 trainable parameters and was trained on a single Tesla V100 for 20 epochs over two days and 19 hours.

\section{Results}
\begin{figure}[ht]
    \centering
    \includegraphics[width=1\textwidth]{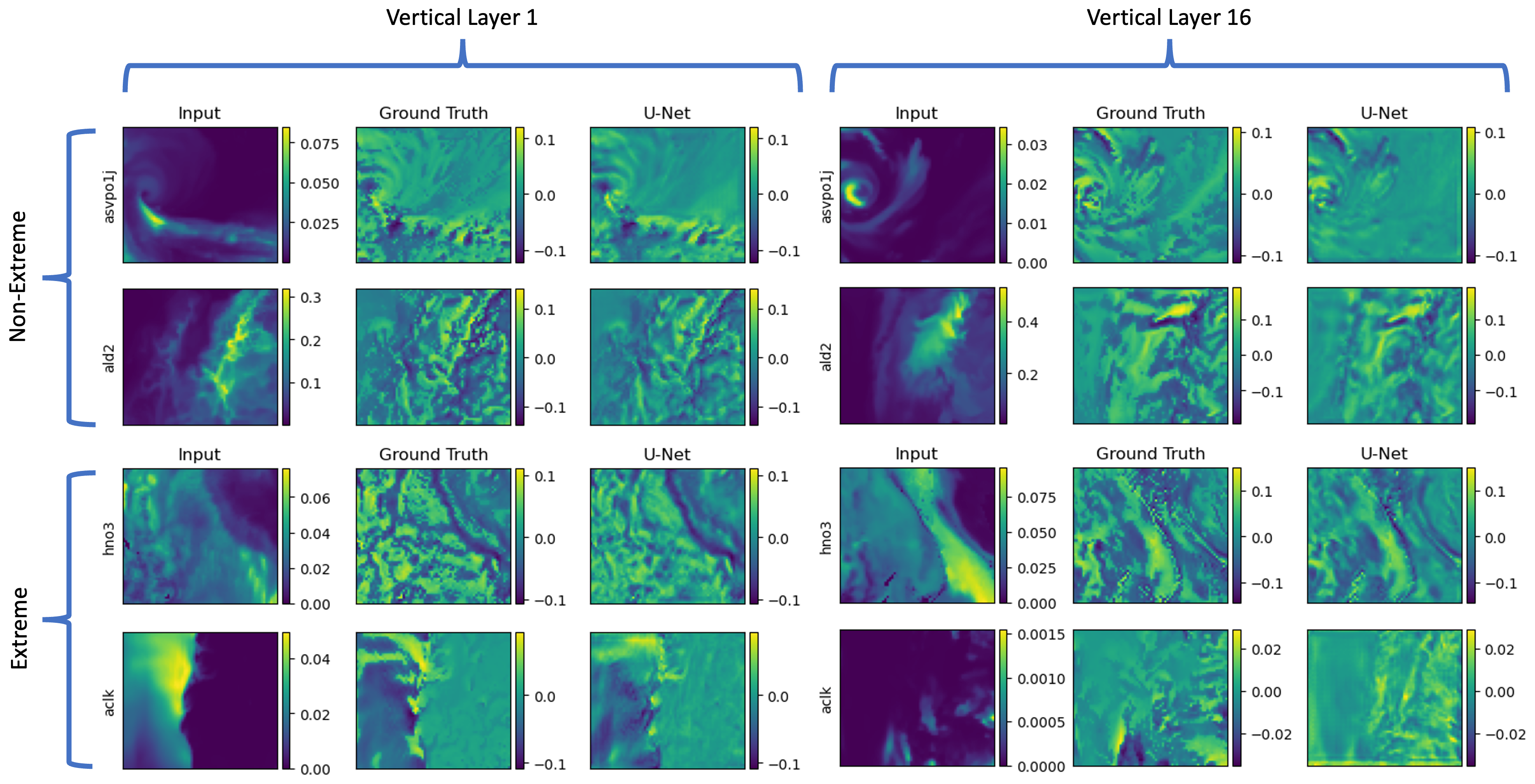}
    \caption{U-Net predictions, where \( Ground \; Truth = (Output - Input)^{1/3}\). The first three columns show patches from vertical layer 1, and the last three columns show patches from vertical layer 16. The first two rows show non-extreme patches, and the last two rows show extreme patches. Species \emph{asvpo1j} and \emph{aclk} are PM2.5 constituents, and \emph{ald2} and \emph{hno3} are contributors to ozone and PM2.5 formation respectively.}
    \label{fig:Difference}
\end{figure}

Figure~\ref{fig:Difference} demonstrates that the U-Net's predictions align well with the ground truth across multiple species, vertical layers, and non-extreme/extreme patches. The RMSE across the entire test dataset, calculated in the min-max normalized space, is 0.0115. The model successfully predicts concentration changes during non-extreme and extreme events with RMSEs of 0.00838 and 0.0129 respectively (Table~\ref{tab:rmses}). Inference on a Tesla V100 takes 4.74 ms for a batch of 32 patches. Extrapolating this time to the entirety of CONUS for all 183 species and 64 layers, the ML approach can produce a prediction in only 2.6 seconds per timestep.

\begin{table}[ht]
  \centering
  \caption{U-Net RMSEs calculated in the min-max normalized space for non-extreme patches, extreme patches, and all patches in the test dataset.}
  \label{tab:rmses}
  \newcolumntype{C}[1]{>{\centering\let\newline\\\arraybackslash\hspace{0pt}}m{#1}}
  \begin{tabular}{|C{2cm}|C{2cm}|C{2cm}|}
    \hline
    Non-Extreme & Extreme & All Data \\
    \hline 
    0.00838 & 0.0129 & 0.0115 \\
    \hline
  \end{tabular}
\end{table}


In an attempt to quantify the model's efficacy in learning the underlying physics, the mass of \emph{asvpo1j} (a PM2.5 constituent) was computed as a preliminary physics-based evaluation metric. The mean percent difference in \emph{asvpo1j} mass, calculated between the U-Net prediction and the UFS-AQ transport module, is 0.0741\%. This demonstrates the potential of the U-Net model in preserving the mass advected by the UFS-AQ model.

\section{Conclusions}
Our model emulates the per-timestep advective transport of atmospheric chemical species. With an overall RMSE of 0.0115, good performance during both extreme and non-extreme AQ events (Table~\ref{tab:rmses}), and estimated prediction time of 2.6 seconds on a single GPU, this ML method exhibits significant potential for integration into the NOAA operational AQ environment. To achieve these results, we utilize data transformations that the underlying transport, governed by the advection equation, is largely invariant under. 

\subsection{Future Work}
In future work, we plan to further explore data transformations, applied on a per-patch basis, which preserve the underlying transport. These include log transformations on the input and output data of the UFS-AQ model, as well as $n$'th root transformations on the residuals. To eliminate boundary artifacts, we will train on larger, overlapping patches. We will also explore adding physics-informed regularization terms to the MSE loss function and further evaluate mass conservation. Ultimately, we aim to develop an ML model which efficiently emulates advective transport over large time-scales for implementation in the UFS-AQ model.


\begin{ack}
This work has been funded by NOAA NA21OAR4310383, SUBAWD003728.
\end{ack}

\printbibliography[title={References}]

\appendix
\section{Appendix}

\begin{figure}[ht]
    \centering
    \includegraphics[width=0.5\textwidth]{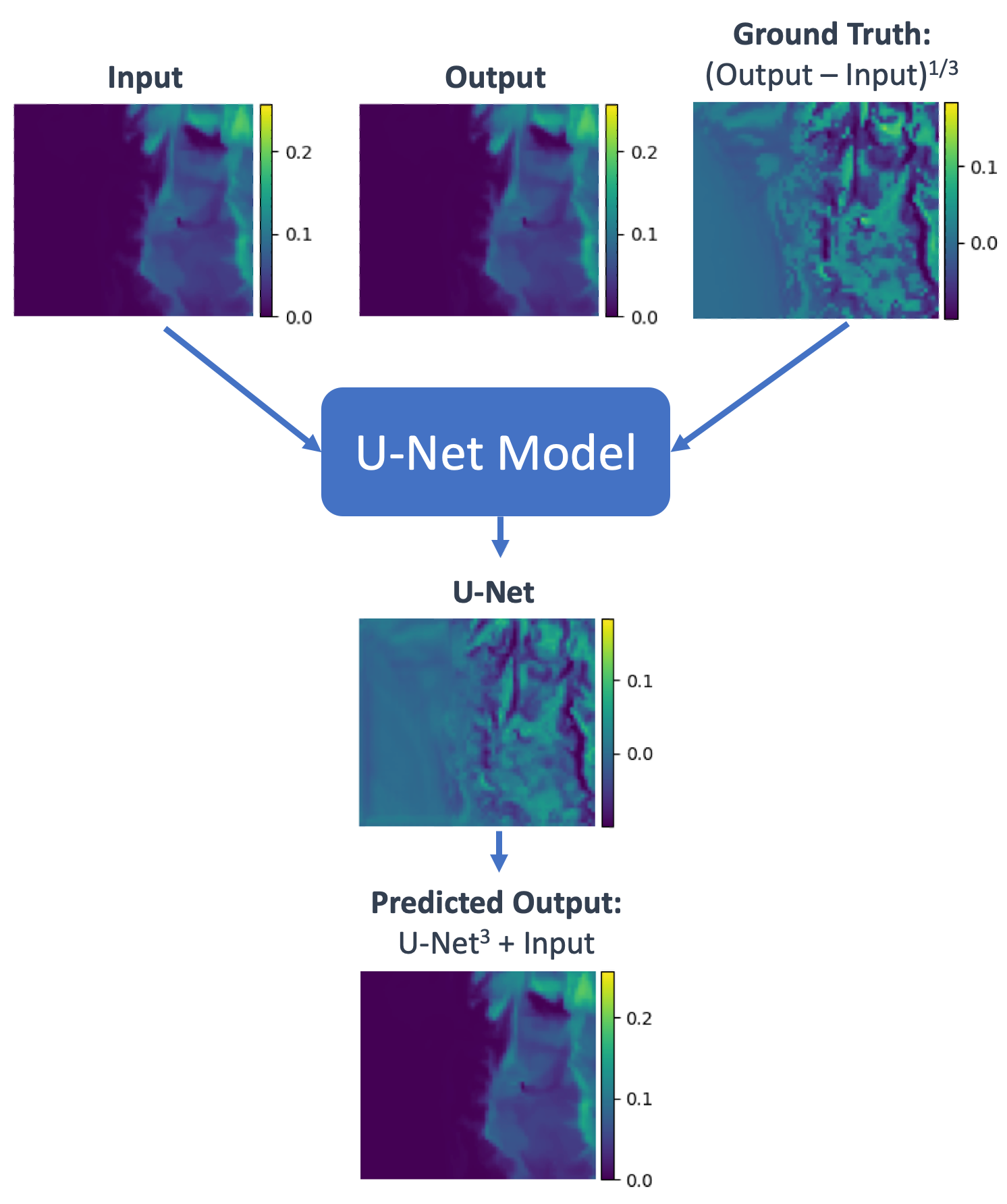}
    \caption{Illustration of the difference learning approach. The U-Net learns the mapping between $Input$ and \((Output - Input)^{1/3}\). The $Predicted \; Output$ is the U-Net's prediction in the min-max normalized space.}
    \label{fig:diagram}
\end{figure}

\begin{figure}[ht]
    \centering
    \includegraphics[width=1\textwidth]{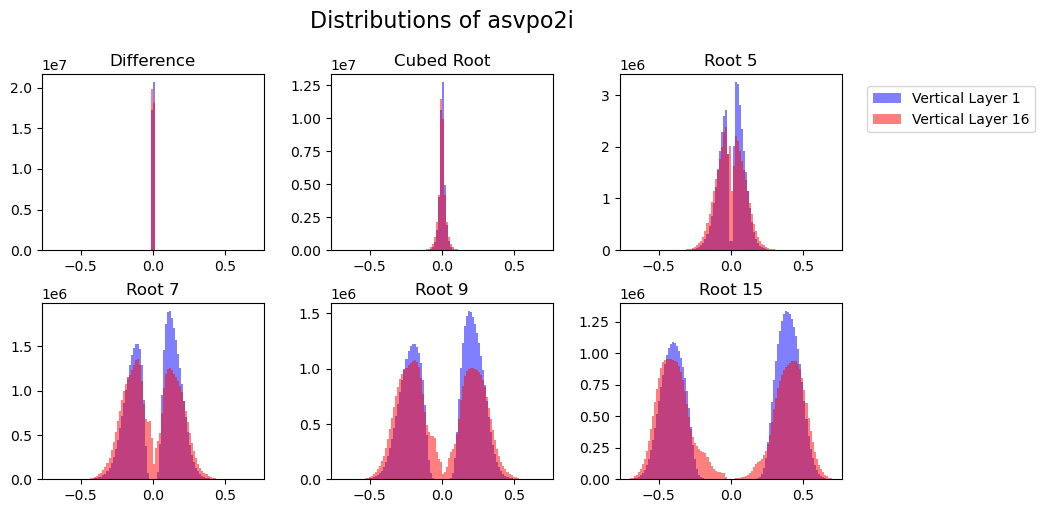}
    \caption{Effects of \((Output - Input)^{1/n}\) transformations on the distribution of \emph{asvpo2i}, a PM2.5 constituent, at vertical levels 1 (blue) and 16 (red). From left to right, top to bottom, $n = 1, 3, 5, 7, 9, 15$.}
    \label{fig:root}
\end{figure}

\end{document}